\documentclass[fleqn,10pt]{wlscirep}
\usepackage[utf8]{inputenc}
\usepackage[T1]{fontenc}
\usepackage{lineno}
\usepackage{subcaption}
\usepackage{caption}
\title{The case for delegated AI autonomy for Human AI teaming in healthcare}

\author[1,*]{Yan Jia}
\author[2,3,+]{Harriet Evans}
\author[1,+]{Zoe Porter}
\author[4]{Simon Graham}
\author[1]{John McDermid}
\author[1,5]{Tom Lawton}
\author[2,6]{David Snead}
\author[1]{Ibrahim Habli}
\affil[1]{Department of Computer Science, University of York, York, UK}
\affil[2]{Histopathology Department, University Hospitals Coventry and Warwickshire NHS Trust, Coventry, UK}
\affil[3]{Warwick Medical School, University of Warwick, Coventry, UK}
\affil[4]{Histofy Ltd, Coventry, UK}
\affil[5]{Improvement Academy, Bradford Teaching Hospitals NHS Foundation Trust, Bradford, UK}
\affil[6]{Department of Computer Science, University of Warwick, Coventry, UK}

\affil[*]{yan.jia@york.ac.uk}

\affil[+]{these authors contributed equally to this work}

\begin{abstract}
In this paper we propose an advanced approach to integrating artificial intelligence (AI) into healthcare: autonomous decision support. This approach allows the AI algorithm to act autonomously for a subset of patient cases whilst serving a supportive role in other subsets of patient cases based on defined delegation criteria. By leveraging the complementary strengths of both humans and AI, it aims to deliver greater overall performance than existing human-AI teaming models. It ensures safe handling of patient cases and potentially reduces clinician review time, whilst being mindful of AI tool limitations. After setting the approach within the context of current human-AI teaming models, we outline the delegation criteria and apply them to a specific AI-based tool used in histopathology. The potential impact of the approach and the regulatory requirements for its successful implementation are then discussed.
\end{abstract}
\begin{document}

\flushbottom
\maketitle
\thispagestyle{empty}

\section*{Introduction}

Artificial intelligence (AI) applications in healthcare are advancing rapidly and their successful implementation depends on effective human-AI teaming \cite{henry2022human}. Currently, the human-AI teaming in healthcare tends to position AI tools in a purely supportive role, so that all clinical decision-making authority remains with human clinicians. There are two main modalities of this approach\cite{barinov2019impact}: sequential clinical workflow, where after initial review by the clinician, the output from the AI-based tool is revealed to the clinician for further consideration before they make a final decision; and concurrent clinical workflow, where the output from the AI-based tool is available from the beginning of the clinician’s decision-making process. These two modalities are shown in Figure 1.

\begin{figure}[!ht]
\centering
\includegraphics[width=0.65\linewidth]{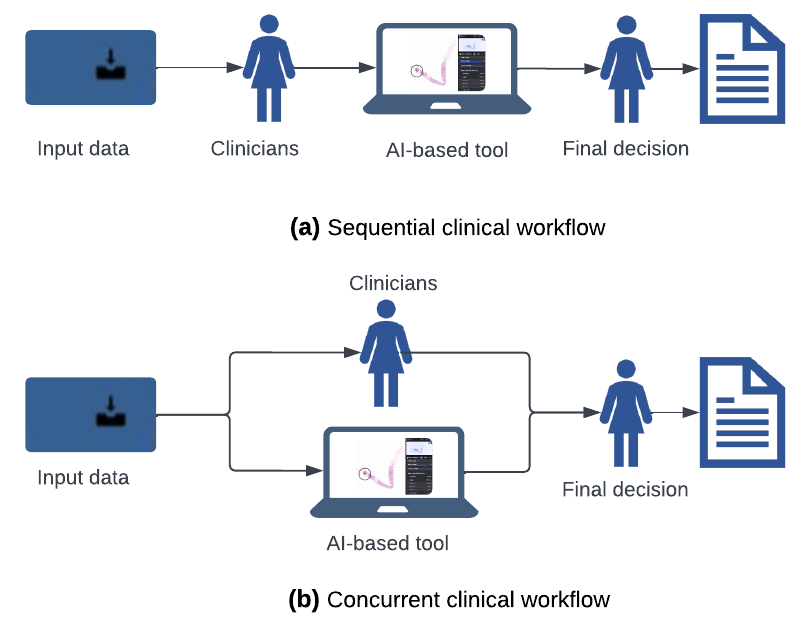}
\caption{Current Human-AI teaming modalities. This figure shows two clinical workflow models for AI-based decision support tools which represents the current approach to integrating AI into clinical decision making processes. The panel (a) represents a sequential clinical workflow, where input data is first processed by clinicians and then passed to the AI-based tool in a linear, step-by-step manner before arriving at the final decision made by the clinician. The panel (b) shows a concurrent clinical workflow, where clinicians and the AI-based tool simultaneously receive and process input data, then the clinician  makes the final decision.}
\label{fig:1}
\end{figure}

The intent is that retaining a human-in-the-loop for all decisions provides patients with a safeguard against potential adverse effects stemming from AI errors and that clinicians will be aided to make better decisions with the support of AI. But there is a paucity of evidence that using AI tools in this way correlates with overall improved performance \cite{vasey2021association} and it might even reduce the sensitivity of diagnosis \cite{lehman2015diagnostic}. Some evidence \cite{gaube2021ai} shows that providing AI-based decision support can lead to unconscious bias in clinician’s decision making processes, e.g. anchoring clinicians to a particular diagnosis or triggering confirmation bias resulting in reduced diagnostic accuracy. This suggests that the current model of keeping AI in a purely supportive role limits the potential of human-AI teaming. Additionally, it restricts the potential efficiency gains of AI in healthcare, since a human is always required in the decision-making process which places demands on stretched healthcare professionals. 
Given that the types of error made by AI and those made by humans are often different \cite{ruamviboonsuk2019deep}(for example, AI is susceptible to artefacts or noise that humans normally would not notice \cite{goodfellow2014explaining}, yet it can also detect tiny signs of cancer which are practically invisible to human eyes \cite{tinycancer}), the ideal of human-AI teaming is to leverage the complementary strengths of human and AI to achieve better overall performance. Against this backdrop, we propose a new approach to human-AI teaming, which we call autonomous decision support. This approach reflects recent developments in this field, but also takes a step further by clarifying the clinical workflow associated with it and articulating an extended set of delegation criteria, making it more practical and comprehensive. 

\section*{Autonomous decision support}

Recently, alternative approaches to the sequential and concurrent clinical workflow modalities in human-AI teaming in healthcare have been proposed. These proposals\cite{dyer2021diagnosis,leibig2022combining,dvijotham2023enhancing,smith2023real} have shown that using AI to autonomously handle a subset of patient cases, guided by the AI’s confidence score in its prediction or recommendation, can improve overall diagnostic performance, e.g. improvements of 2.6\% in sensitivity and 1.0\% in specificity have been achieved compared with unaided radiologists \cite{leibig2022combining}. This is particularly significant given that there is usually a trade-off between these desiderata, with increased sensitivity coming at the expense of decreased specificity, and vice versa \cite{florkowski2008sensitivity}. However, there are limitations in these approaches. For example, Dvijotham et al \cite{dvijotham2023enhancing} have proposed a complementarity-driven deferral to clinical workflow (CoDoc) approach, which either relies solely on AI predictions if the AI is confident or defers entirely to human decisions when the AI is not confident. But this overlooks the fact that not all decisions carry the same criticality, therefore some decisions still require human intervention even if the AI is confident in its predictions. Smith et al \cite{smith2023real} have proposed to use AI algorithms autonomously reporting the normal chest radiographs that the AI predicted and has high confidence in, i.e. ‘high confidence normal’(HCN), while directing the remaining cases to the normal clinical workflow in the hospital. Compared to the CoDoc approach, this approach suggested that only the normal radiographs should be handled autonomously. However, it fails to leverage the value of AI in the abnormal cases, where it might identify clinically significant lesions that might be missed by humans. Similarly, Leibig et al \cite{leibig2022combining} have proposed a decision-referral approach for autonomously reporting the HCN, and refer the rest to the normal clinical workflow without indicating the AI predictions - with the addition that a warning is displayed if clinicians miss a lesion that the AI is confident in. This approach is the most similar to our proposal, however it fails to recognise that other important factors, beyond AI confidence scores and normal or abnormal predictions, should be considered when directing patient cases. The clinical context is fundamental to all medical decision making, including interpretive tasks required in radiology and cellular pathology and so should be part of determining AI tool deployment, in order to ensure the maximum benefit and minimal risk.

\begin{figure}[!ht]
\centering
\includegraphics[width=0.8\linewidth]{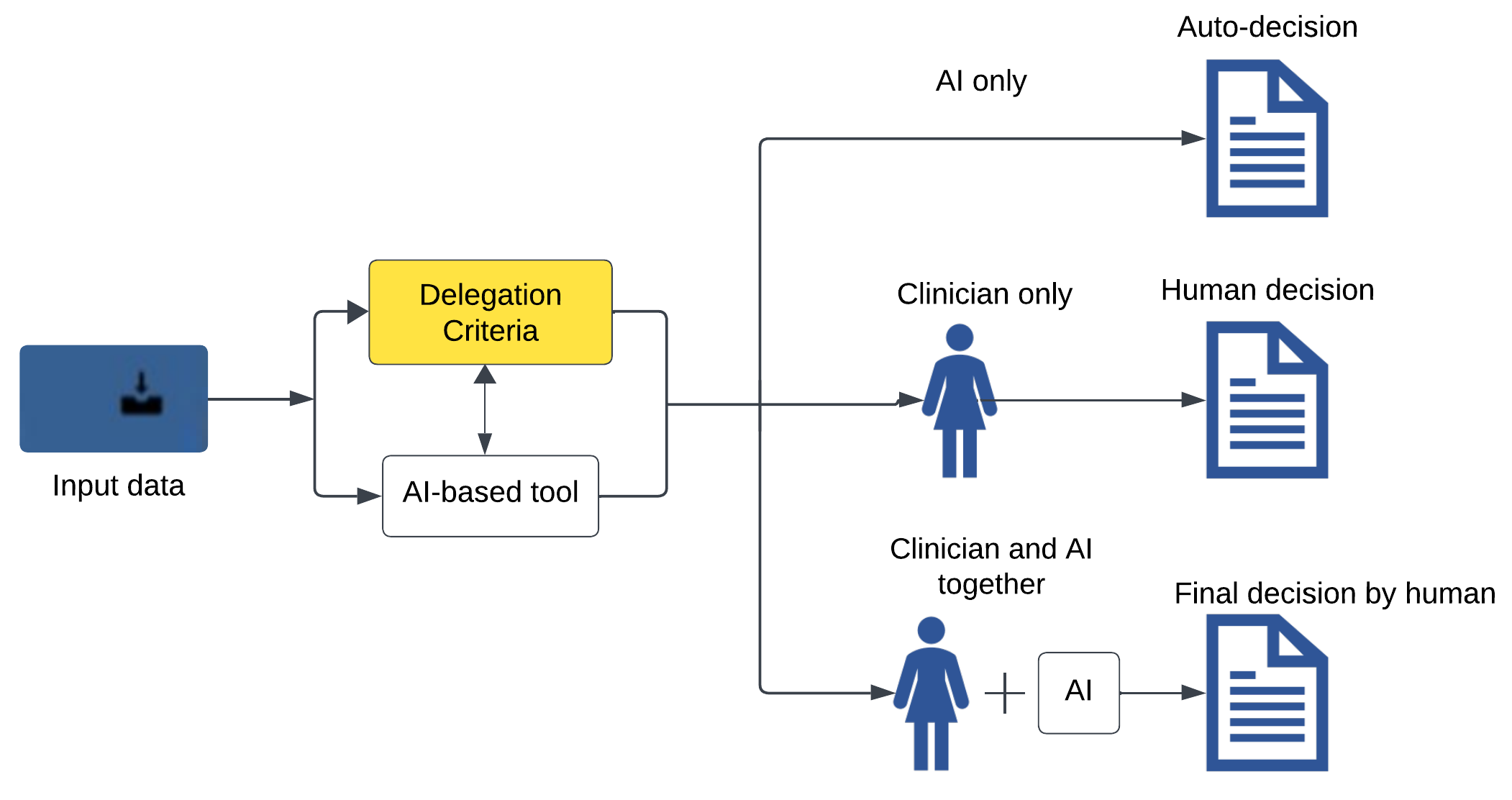}
\caption{The clinical workflow for an autonomous decision support tool. This figure shows our approach for integrating AI in healthcare, where it has a delegation criteria, which interacts with the AI-based tool, to direct the input patient data to three distinctive pathways, ``AI only'', ``Clinician only'', and ``Clinician and AI together''. ``AI only'' pathway represents the ``autonomous'' element in our approach and ``Clinician and AI together'' pathway represents the ``support'' element in our approach, hence the name ``autonomous decision support'' approach.}
\label{fig:2}
\end{figure}

Therefore, we propose an autonomous decision support approach, as shown in Figure 2 where the patient and case data is utilised to \textcolor{black}{determine the most appropriate usage for the AI-based tool}. Rather than automatically giving the AI prediction, delegation criteria are utilised to direct the case into one of three distinct pathways: 
\begin{enumerate}
    \item AI only: the AI prediction is taken as the final decision without human intervention; 
    \item Clinician only: the AI does not make a prediction and refers the case to a human; 
    \item Clinician and AI together: the AI prediction is revealed to support the clinician, who then makes the final decision, akin to existing human-AI teaming modalities (see Figure 1). 
\end{enumerate}

Compared to existing human-AI teaming models in Figure 1, our approach enhances the third pathway by informing the clinician of the AI’s prediction rationale, thereby providing an efficient transfer of information which the clinician can accept, reject or modify as they deem appropriate. For instance, instead of displaying AI predictions for all patient cases, the system could only present confident abnormal predictions to the clinician, prompting a more careful review. Overall, our autonomous decision support approach leverages the benefits of AI in both supportive and autonomous roles, enhancing overall decision-making and patient safety. It ensures that automation is only delivered for cases for which it is safe to rely on AI (e.g. high confidence recognition of normal cases) and flags cases for clinician review when \textcolor{black}{the prediction carries more criticality, e.g. abnormal predictions, even when the AI is confident}. This allows safe handling of the patient cases whilst allowing clinicians to potentially reduce review time and be mindful of AI limitations. Additionally, our approach also allows certain cases to be handled solely by clinicians when the input is beyond the AI’s capability, e.g. out of scope cases, \textcolor{black}{or when the AI is less reliable, e.g. if a patient's clinical context indicates AI prediction of normal is not to be relied upon.}

The design of the delegation criteria, which directs patient cases into these three distinct pathways, is the core of the autonomous decision support approach. In the work cited above, the delegation criteria mainly rely on the utilisation of AI confidence scores and its predictions. In this paper, we present a more comprehensive set of factors to inform the delegation criteria, including the \textit{types of tasks}, \textit{clinical context}, \textit{criticality of the decision}, \textit{AI confidence score}, and the \textit{failure modes of both humans and AI}, as shown in Table 1. \textcolor{black}{The factors presented here are intended to be general, so that they can be applied widely. Therefore, it is necessary to operationalise the delegation criteria for a specific application. This necessitates collaboration within a multidisciplinary team involving AI developers and clinicians, who between them have the respective expertise to support the development and implementation of the delegation criteria. This multidisciplinary collaboration is becoming more widely recognised with medical devices companies employing clinicians working alongside AI developers, but is not yet common.}

\subsection*{Delegation Criteria}

\begin{table}[!ht]
% \usepackage{graphicx}
\resizebox{\textwidth}{!}{%
\begin{tabular}{|l|l|l|}
\hline
\textbf{Factors} & \textbf{Description} & \textbf{\textcolor{black}{Design questions}} \\ \hline
Types of tasks & \begin{tabular}[c]{@{}l@{}}The type of task that the AI \\ is going to be applied to, e.g. \\ image classification, treatment \\ recommendation, etc.\end{tabular} & \begin{tabular}[c]{@{}l@{}}Is the task to which the AI will be applied suitable for \\ implementing the delegation criteria? If so, define the \\ exact type of input data, the output of the AI tool and \\ the patient group for which the AI tool will be used.\end{tabular} \\ \hline
Clinical context & \begin{tabular}[c]{@{}l@{}}Factors encompassing the \\ patient's medical history, \\ current symptoms, results of \\ other investigations and \\ treatment history.\end{tabular} & \begin{tabular}[c]{@{}l@{}}What clinical context is the AI tool going to be used \\ in including the current clinical pathway? Are there any \\ explicit factors in the clinical context that cannot be \\ accommodated by the AI? If so, consider incorporating \\ these factors into the delegation criteria.\end{tabular} \\ \hline
Criticality of the decision & \begin{tabular}[c]{@{}l@{}}The importance or \\ significance of a decision \\ in a given clinical context.\end{tabular} & \begin{tabular}[c]{@{}l@{}}How much impact does each output of the AI tool have on \\ patient care? If an output of the AI tool carries greater\\ significance, then involving human would be necessary.\end{tabular} \\ \hline
AI confidence score & \begin{tabular}[c]{@{}l@{}}Quantification of the \\ uncertainty in an AI output\end{tabular} & \begin{tabular}[c]{@{}l@{}}Can a confidence score be utilised to indicate the reliability \\ of the AI prediction, e.g. through a validation study? If so, \\ consider combining the use of the confidence score and \\ AI prediction in the delegation criteria.\end{tabular} \\ \hline
Failure modes of AI & \begin{tabular}[c]{@{}l@{}}The scenarios where the AI \\ often gives erroneous output \\ even when the input is within \\ scope\end{tabular} & \begin{tabular}[c]{@{}l@{}}Are there any known failure modes of the AI tool, e.g. \\ through validation study, which are not easy to resolve \\ by further algorithm training? If so, consider using the \\ failure modes of AI to direct these types of patient cases \\ to humans.\end{tabular} \\ \hline
Failure modes of humans & \begin{tabular}[c]{@{}l@{}}The scenarios where humans \\ often makes erroneous \\ judgement\end{tabular} & \begin{tabular}[c]{@{}l@{}}Are there any known failure modes of humans in this \\ clinical context? If so, consider using the AI tool to \\ support the human in these types of cases.\end{tabular} \\ \hline
\end{tabular}%
}
\caption{Factors that influence the design of Delegation Criteria }
\label{tab:my-table}
\end{table}

\textcolor{black}{In table 1, we have shown the detailed descriptions for each factor, accompanied by a set of design questions tailored for AI tool developers,  clinical team and other relevant experts to discuss in order to facilitate the design of the delegation criteria. Overall, there are three important steps to consider when developing delegation criteria for the AI tool.} 

\begin{enumerate}
    \item \textcolor{black}{The first crucial step is to evaluate the \textbf{\textit{type of task}}. This initial assessment determines whether a clinical task is suitable for implementing the delegation criteria or not. Tasks with less context dependency, such as radiology and pathology, which primarily involve image analysis and laboratory results, tend to be more suitable to implement the delegation criteria than tasks in an ICU setting, where patient conditions can be highly variable and require real time adjustments.}
    \item \textcolor{black}{If the task is determined to be suitable for implementing the delegation criteria, then it is important to clearly delineate the \textbf{\textit{type of task}} an AI tool is performing to ensure correct use and safety. This involves precise specification of: (i) the exact type of input data the AI system can process, for example, within pathology this involves defining the anatomical site, type of specimen, type of tissue stain. (ii) what output the AI system can produce, e.g. binary output or multi-class classification (iii) which patient group for which the AI tool can safely operate, e.g. adult. 
    The precise definition of the type of input data and the task to be performed ensures that wherever possible AI is utilised in decision making, but out of scope scenarios will be identified for human intervention.}
    \item \textcolor{black}{The third step involves considering the other factors, i.e. \textbf{\textit{Clinical context, Criticality of the decision, AI confidence score, Failure modes of AI and human}} to further define how the patient cases are routed to the three pathway outlined in Figure 2. These five factors are not hierarchical; instead, they should be considered simultaneously to weave into the design for the delegation criteria. Further, each factor may need to revisited multiple times during the process. In the following, we illustrate these five factors in detail.}
\end{enumerate}

%Clear delineation of the \textbf{type of task} an AI tool is performing is essential to ensure correct use and safety. The precise definition of the type of input data and the task to be performed ensures that wherever possible AI is utilised in decision making, but out of scope scenarios will be identified for human intervention. The type of task is largely shaped by the development of the tool and the data used to train the model. For example, within pathology this involves defining the anatomical site, type of specimen, type of tissue stain and patient group for which the AI tool can safely operate.

\textbf{Clinical context} is crucial when making decisions in healthcare. It encompasses the patient’s medical history, current symptoms, results of other investigations and treatment history. This contextual view allows clinicians to make informed decisions that are tailored to the individual patient. For instance, when a pathologist or radiologist is reviewing a patient’s specimen or imaging, they consider whether their diagnosis would match the clinical context, or if the patient is more at risk of certain conditions due to their medical history. Despite advances in AI technology, current systems lack the ability to  integrate and interpret this nuanced clinical context. Therefore, consideration of the clinical context in which the AI tool is used should be reflected in the design of the delegation criteria. \textcolor{black}{Clinical context not only helps to determine whether a patient case is suitable for passing to the AI tool, but also provides insight into the reliability of the AI predictions}.

When considering the \textbf{criticality of decisions}, it is important to recognise that even within the same task, different outputs can have vastly different patient implications. For instance, when examining colonic (large bowel) biopsies,  abnormalities cover an enormous range of conditions from minor changes requiring no further intervention through to life threatening disease such as cancer. This range of likely patient impact illustrates the need for careful consideration of decision criticality in designing delegation criteria. \textcolor{black}{Ideally, the most critical decisions should combine AI and clinician perspectives}.

For the \textbf{AI confidence score}, it is important to quantify uncertainty from both known and unknown data distributions (also known as out-of-distribution where the input data is from a significantly different distribution compared to the training data). Ideally, an AI confidence score should express high uncertainty in cases of incorrect predictions and out-of-distribution inputs. However, in reality, neural networks are known to produce overly confident predictions \cite{nguyen2015deep,lakshminarayanan2017simple}. There are various methods that can be employed to calculate a confidence score, ranging from basic techniques such as utilising probabilities from softmax distributions to more advanced approaches such as Bayesian methods \cite{blundell2015weight, gal2016dropout} or ensemble methods \cite{kompa2021second}. Despite extensive research, there is no consensus on the optimal method for determining confidence scores \cite{gawlikowski2023survey}. Studies \cite{hendrycks2016baseline} have shown that while the AI tends to produce over confident predictions, correctly classified examples tend to exhibit higher maximum softmax probabilities compared to erroneously classified samples and out-of-distribution samples, which could serve as a baseline for confidence scoring. Nonetheless, solely relying on the AI confidence in the delegation criteria can be problematic as these scores may not accurately reflect the real world conditions. Thus, it's crucial to consider additional factors to improve the reliability of the delegation criteria.

In terms of \textbf{failure modes of AI}, more studies are beginning to investigate the nature of errors made by AI systems. For instance, recent studies on a prostate cancer screening tool reported their false negative errors, highlighting that the tool struggled with variants of adenocarcinoma such as foamy gland carcinoma, adenocarcinoma with hormone deprivation therapy effects, adenocarcinoma with an atrophic appearance, low grade adenocarcinoma or isolated foci of perineural invasion \cite{da2021independent,perincheri2021independent,raciti2020novel}. This knowledge of AI error patterns can guide the design of the delegation criteria (if not resolved by further algorithm training), prompting the delegation of cases prone to AI errors to human clinicians. In order to do this, there should be emphasis on reporting AI errors in clinically relevant terms rather than solely technical metrics \cite{evans2024understanding}, e.g. false negatives or positives. 

Categorisation of the \textbf{failure modes of humans} may be difficult due to limited published work and the inherent variability and complexity involved in categorising human errors. Nonetheless, \textcolor{black}{where available, studies should be utilised to provide insight into existing human errors}. For instance, in the examination of colon biopsies, one study \cite{azam2024digital} identified that the two most frequent areas prone to errors were the identification of the type of serrated polyp and the grading of dysplasia within a polyp. By pinpointing these specific areas of vulnerability, AI predictions can potentially complement human expertise and ultimately improve diagnostic accuracy and patient care.

Identifying these factors represents the initial step in designing the delegation criteria. \textcolor{black}{Implementing these criteria requires careful consideration both in the design of the AI tool (such as ensuring the tool can filter out data that is not part of its defined task) and how it is utilised in practice. This requires definition of the intended use case for which the healthcare system allows a tool to operate and how data flow is managed in the pathway (such as ensuring that cases with a low AI tool confidence are automatically sent to a clinician to review).} Importantly, the delegation criteria must evolve in tandem with the AI system itself; any updates to the AI system should prompt a review and corresponding update of the delegation criteria. In the following section, we illustrate how our approach can be applied to an AI tool in histopathology.  

\section*{Applying autonomous decision support in histopathology: COBIx}
The colon can be affected by a wide range of inflammatory and neoplastic diseases, diagnosed primarily by endoscopic examination and histological sampling. The increased volume of endoscopic biopsies, as well as an increasing complexity of reporting has led to an extra demand on already struggling pathology services. Currently, large bowl biopsies account for 10\% of all tissue sample requests in laboratories with approximately 30-40\% of these samples being normal. Therefore, implementing an autonomous decision support approach that can automatically report some normal biopsies and support pathologists with other cases would be highly beneficial. To this end, an AI-based tool has been developed to assist in the analysis of Colon and Rectal Endoscopic Biopsy (COBIx). Details on the development of the COBIx algorithm can be found in this paper \cite{graham2023screening}. In the following, we illustrate how application of the delegation criteria could determine the most appropriate pathway for COBIx use: AI only, pathologist only, and pathologist and AI (matching Figure 2).

Figure 3 starts with the \textbf{type of task}. COBIx has been trained to classify haematoxylin and eosin (H\&E) histopathology whole slide images (WSI) from colon and rectal biopsy for adult patients, which defines the correct WSI to be processed by COBIx.

\begin{figure}[!ht]
\centering
\includegraphics[width=0.8\linewidth]{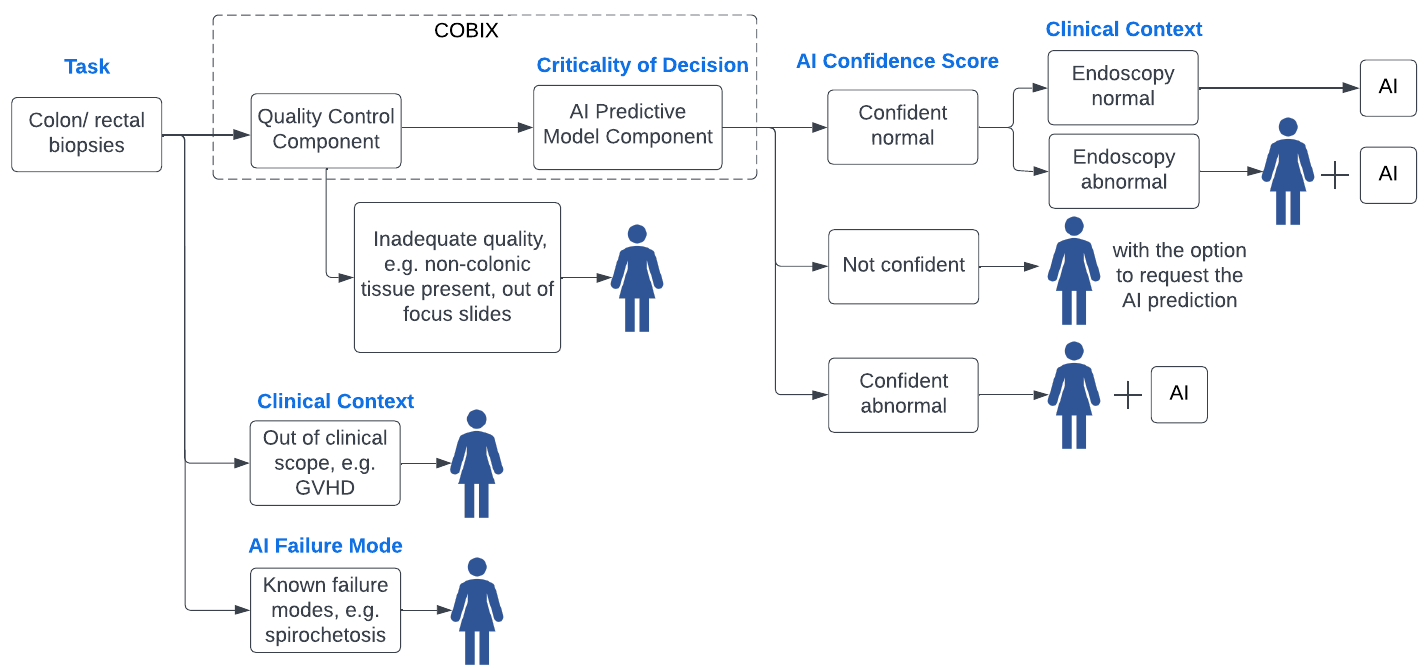}
\caption{The potential clinical workflow for using COBIx as an autonomous decision support tool. COBIx is a published multi-class AI-based tool for analysing colon and rectal (large bowel) endoscopic biopsies. It produces five possible outputs with one normal category and four abnormal categories, i.e. Neoplastic urgent, Neoplastic non-urgent, Non-neoplastic urgent, Non-neoplastic non-urgent. This workflow shows how Clinical context, AI failure mode, Criticality of decision, and AI confidence score can be incorporated into designing the delegation criteria to safely direct the biopsies into three pathways: ``AI only'', ``Clinician only'' and ``Clinician and AI together''.}
\label{fig:3}
\end{figure}

The COBIx assessment of colonic WSI is a multi-step process. First, a quality control step ensures the slides are suitable to be processed by the algorithm. It excludes samples containing non-colonic tissue, e.g. adjacent small bowel (ileal) or squamous (anal) mucosa in the gastrointestinal tract, and those with inadequate colonic tissue to make a diagnosis. Further, poor quality WSIs, such as those that are out of focus or folded, where quality hinders the ability of COBIx to perform its task, will be excluded and directed to a pathologist. 

Next, a segmentation step, trained on large annotated datasets, detects various important structural and cellular regions in the WSI, such as glands, lumen and different nuclei. This segmentation facilitates the identification of histological features used by human pathologists to recognise disease or normality. Then, these  histological features identified on a WSI are fed into a neural network to predict the classification. COBIx classifies biopsies into normal or one of four abnormal categories based on the features present, i.e. Normal, Neoplastic urgent, Neoplastic non-urgent, Non-neoplastic urgent, Non-neoplastic non-urgent, thus \textbf{Criticality of decision} is built into COBIx. It is vital that malignant conditions such as adenocarcinoma, and pre-malignant entities such as high-grade dysplasia, are not missed. Therefore even a small area of these on a WSI is classified as abnormal and must be seen by a pathologist. This use of harsh cut-off criteria for certain high-risk features helps ensure critical decisions are made by pathologists.

Further, COBIx's output needs to be calibrated, so that the output score accurately reflects the \textbf{model's confidence}. For example, a score of 0.5 suggests the model is correct 50\% of the time, indicating low confidence, while a score of 0.9 indicates the model is correct 90\% of the time, reflecting high confidence. To reduce the likelihood of incorrectly screened cases, we can set a threshold to consider only slides with confident predictions. This threshold can be determined by analysing the reduction in the screening error rate. For confident normal cases, an automated report could potentially be issued, without pathologist review, but confident abnormal cases would be reviewed by a pathologist with the support from AI (with the categorisation of different abnormal cases by COBIx allowing cases to be triaged based on urgency). Not confident cases would be reviewed by pathologists with the option to request the AI prediction.

The \textbf{clinical context} is considered at multiple stages in the COBIx processing pipeline. First, despite extensive training, there will be entities COBIx has not been trained to detect due to the rarity in the population. In such scenarios it is inappropriate to use COBIx and instead a pathologist should make the diagnosis. One such example is graft versus host disease (GVHD), a condition only seen following allogenic stem cell transplant or occasionally following solid organ transplant, so knowledge of the patient’s medical history is vital. A patient history of transplantations and/or a clinical suspicion of this entity can thus ensure this specimen is reviewed by a pathologist.

A second, and more fundamental, way in which the \textbf{clinical context} is relevant is the endoscopic information. All colon biopsies are taken during endoscopy, during which the endoscopist can visualise abnormalities in the colon, providing highly relevant insight into the likely disease process. This endoscopic information forms part of a key delegation step in the use of COBIx. If COBIx classifies a specimen as confident normal, and the endoscopy is also normal, this provides strong support that the colon is actually normal and so AI alone can make this diagnosis. However, if COBIx reports the biopsy as confident normal but the endoscopy report is abnormal, this contradiction prompts suspicion that the COBIx classification could be incorrect therefore a pathologist must review the case. The use of endoscopic information specifically for this filtration of slides classified as confident normal by COBIx also reflects \textbf{criticality of decision}, because false negative errors - potentially resulting in disease progression and morbidity - would have the greatest clinical significance for patients, making it critical to minimise false negatives. 

Knowledge of COBIx \textbf{AI failure modes} could further enable tailored use of the tool. For example, if COBIx struggled to detect an entity such as spirochetosis (which is associated with very little other changes in the colonic tissue so would be otherwise undetected), then clinical suspicion of this entity can direct the case towards a pathologist.

\textbf{Human failure modes} are less relevant as delegation criteria in the current proposed use of COBIx because the scenarios where pathologists struggle are variable and poorly defined. Additionally, COBIx may encounter similar challenges in identifying these cases.

This illustrates how delegation criteria can help ensure the safe and efficient use of an autonomous decision support tool. They enable the use of COBIx for independent diagnosis when appropriate, reducing pathologist workload and minimising delays in reporting. At the same time, they ensure that colon biopsies with abnormalities or out of clinical scope, or in patients with endoscopic abnormalities, are safely reviewed by pathologists.

\section*{Potential impact of the autonomous decision support approach}
Rather than reduce the burden on an already overworked clinician, the purely supportive approaches place an additional task on clinicians, requiring them to act as a sense check on the AI at all times. By contrast, our autonomous decision support approach only imposes this additional task on the clinician in a subset of cases, in accordance with the delegation criteria. Therefore, if implemented, it will allow clinicians to shift their focus to complex cases, meaning all patients receive their results sooner, alleviating any stress from awaiting a diagnosis and allowing quicker initiation of treatment where required. 

This holds great promise for a safer and more efficient healthcare system. Even so, careful evaluation is required. The cases which are autonomously reported by the AI are likely to be those which are normally dealt with most quickly by clinicians. Reporting a reduction in the proportion of cases needing review as if it were an equal reduction in workload is misleading, as the more complex cases left behind are likely to be the ones taking the majority of the clinicians’ time \cite{park2023ai}. Further, we should also consider the impact of autonomous decision support on clinicians over time. Clinicians traditionally develop their expertise starting from simpler cases, so if this approach becomes established, there may be too few simple or normal cases from which to learn and develop their judgement. This is likely both to affect their perception of what is normal and alter their beliefs about the base-rate of normality in the population, with the possible consequence of over-diagnosis. Ensuring that clinician workload, training and understanding is not adversely impacted by the autonomous decision support approach should be at the forefront of its wider evaluation.

\textcolor{black}{An ethical concern arises when an AI tool is in a purely supportive role (whether in the sequential or concurrent clinical workflow model) and where the clinician is expected to assess the validity or appropriateness of the AI's output in order to make the final decision.} Thus, there is a concern that clinicians may end up being unfairly held liable for acting on an AI prediction or implementing an AI recommendation that turns out to be wrong. \textcolor{black}{This has been referred to as clinicians being ‘liability sinks’ for AI \cite{lawton2024clinicians}. The term ‘liability sink’ refers to when one frontline individual (in this context, the clinician) absorbs all of the liability for a harmful outcome in which an AI system was a major cause. The problem is not clinicians being held liable, but that clinicians are being held liable when they have insufficient understanding of how the AI tool reaches its output to be able to evaluate those outputs effectively. }

\textcolor{black}{The autonomous decision support approach proposed here helps to address this liability sink issue because} the clinician is no longer placed in a position of acting as a safeguard on the machine in all cases. In the first pathway, \textcolor{black}{referred to as ‘AI only’ in Figure \ref{fig:2},} the cases are autonomously processed. \textcolor{black}{By removing the clinician as intermediary, this pathway} could shield the frontline clinician from being a liability sink. However,it should also be recognised that clinicians in the medical device company, auditing the patient cases reported by the AI, could face the liability sink problem. In the second pathway, \textcolor{black}{referred to as ‘clinician only’,} the clinician decides without any AI recommendation at all, which is the same as the traditional practice. \textcolor{black}{As such, they would not be absorbing liability for faults or errors of the AI tool. And in the third pathway, referred to as ‘clinician and AI together’,} which is the same as the existing human-AI teaming modality, but the clinician at least has more information about the AI tool's output, \textcolor{black}{e.g. only confident abnormal predictions from the AI are shown to the clinician prompting a more careful review}, therefore the clinician is able to make more informed decisions, which should \textcolor{black}{limit the degree to which they absorb liability unfairly due to a lack of understanding.} However, there is still a need to evaluate and update how liability should be distributed across the wider healthcare system \cite{smith2020artificial, lawton2024clinicians}.

\section*{Establishing new regulatory requirements}

Autonomous decision support marks an advance in the design and use of AI systems that healthcare regulators, e.g. the U.S. Food and Drug Administration (FDA) \cite{fdaapprove}, currently approve and new regulatory requirements need to be established to unlock its benefits. \textcolor{black}{Currently approved AI/ML based medical devices are intended to be used in a purely supportive role, either in a sequential or concurrent clinical workflow. No devices have been approved for autonomous use and there is a need to establish new regulatory requirements specifically for AI tools intended for such use. Further, the design of the delegation criteria, which requires a multidisciplinary team to cover the different areas of expertise,  might be possible within the medical device company in the development phases, but it is more likely to require collaboration between the medical device company and hospitals that use the AI tool, during the deployment phases. Therefore any new regulations need to allow such flexibility when clearing the medical devices. This potentially can be facilitated by the recent draft guidance from the FDA for pre-determined change control plans for medical devices \cite{fdaguidance}, which describes what modifications can be made to medical devices and how they can be assessed after approval.}\textcolor{black}{Further, this will also likely affect both technical and clinical validity. }
%Development of the delegation criteria requires medical device manufacturers to invest additional effort in understanding their AI systems, not to mention the adaption of the delegation criteria as the systems evolve. Beyond merely assessing the overall technical metrics, they must discern the specific scenarios where performance excels and where it may falter in order to develop safe delegation criteria. Moreover, the manufactures need to have a good audit system which allows hospitals to trace back and evaluate every case handled by the AI systems. This could include new requirements specifically related to the delegation criteria along with the general requirements for the AI systems, e.g. technical validity and clinical validity of the product. 

In terms of technical validity, one key issue is to evaluate the calibration between AI confidence scores and the accuracy of the AI model. This calibration should demonstrate that as AI confidence increases, the accuracy of predictions also improves, ideally showcasing a linear relationship between confidence level and prediction accuracy, ensuring a consistent and proportional increase in accuracy with higher confidence score. 

In terms of clinical validity, while prospective and retrospective studies are commonly used, evaluating the longer-term impact of this approach necessitates a novel methodology. This involves finding ways to clinically follow up patients, ensuring that the approach does not systematically disadvantage any particular demographic group, while preserving patient privacy. Further, there also needs to be ways to evaluate the impact on clinicians learning ability without turning this into unconstructive criticism of individual clinician performance. By addressing both technical and clinical validity through comprehensive evaluation, we can foster greater confidence in this new approach to integrating AI-driven technologies in healthcare systems. 

\section*{Conclusion}

Introducing AI in healthcare necessitates investment in training and infrastructure, but our autonomous decision support approach can minimise extra demands by leveraging available clinical information for the design of the delegation criteria. Further, the transparency and adaptability of the delegation criteria would be necessary to support the adoption and regulatory approval of this approach. Finally, our approach further emphasises the importance of a multidisciplinary team in AI tool development, including clinicians, AI developers, ethicists and other relevant stakeholders.

%\noindent LaTeX formats citations and references automatically using the bibliography records in your .bib file, which you can edit via the project menu. Use the cite command for an inline citation, e.g.  \cite{Hao:gidmaps:2014}.

%For data citations of datasets uploaded to e.g. \emph{figshare}, please use the \verb|howpublished| option in the bib entry to specify the platform and the link, as in the \verb|Hao:gidmaps:2014| example in the sample bibliography file.

\section*{Acknowledgement}

This work was supported by the MPS Foundation Grant Programme (2022-0000000206: Shared CAIRE – Shared Care AI Role Evaluation), the Engineering and Physical Sciences Research Council (EP/W011239/1: Assuring Responsibility for Trustworthy Autonomous Systems), the National Institute for Health and Care Research (AI\_AWARD02688: COBIx: Multi-site validation of automated AI tool for screening of large bowel endoscopic biopsy slides) and the Centre for Assuring Autonomy (funded by the Lloyd's Register Foundation and the University of York). 

\section*{Author contributions statement}
%Must include all authors, identified by initials, for example:
Y.J. conceived the idea and contributed to the design and structure of the manuscript. J.M., T.L., and I.H contributed to the conception of the idea. Z.P. contributed to the analysis of the potential impact of the approach. H.E., S.G., and D.S. contributed to the histopathology COBIx example. All authors contributed to the writing and reviewing of the manuscript. 

\section*{Competing interest}

Prof Tom Lawton reports he is the Head of clinical AI for Bradford Teaching Hospitals NHS Foundation Trust. Prof David Snead reports he is co-founder, shareholder and director of Histofy Ltd and has received honoraria from Astra Zeneca and Oliver Wyman LLC. Simon Graham reports he is co-founder and shareholder of Histofy Ltd. Harriet Evans reports working ad hoc sessions for Histofy Ltd. The remaining authors declare no competing interests.

\bibliography{reference}

\begin{thebibliography}{10}
\urlstyle{rm}
\expandafter\ifx\csname url\endcsname\relax
  \def\url#1{\texttt{#1}}\fi
\expandafter\ifx\csname urlprefix\endcsname\relax\def\urlprefix{URL }\fi
\expandafter\ifx\csname doiprefix\endcsname\relax\def\doiprefix{DOI: }\fi
\providecommand{\bibinfo}[2]{#2}
\providecommand{\eprint}[2][]{\url{#2}}

\bibitem{henry2022human}
\bibinfo{author}{Henry, K.~E.} \emph{et~al.}
\newblock \bibinfo{journal}{\bibinfo{title}{Human--machine teaming is key to ai adoption: clinicians’ experiences with a deployed machine learning system}}.
\newblock {\emph{\JournalTitle{NPJ digital medicine}}} \textbf{\bibinfo{volume}{5}}, \bibinfo{pages}{97} (\bibinfo{year}{2022}).

\bibitem{barinov2019impact}
\bibinfo{author}{Barinov, L.} \emph{et~al.}
\newblock \bibinfo{journal}{\bibinfo{title}{Impact of data presentation on physician performance utilizing artificial intelligence-based computer-aided diagnosis and decision support systems}}.
\newblock {\emph{\JournalTitle{Journal of Digital Imaging}}} \textbf{\bibinfo{volume}{32}}, \bibinfo{pages}{408--416} (\bibinfo{year}{2019}).

\bibitem{vasey2021association}
\bibinfo{author}{Vasey, B.} \emph{et~al.}
\newblock \bibinfo{journal}{\bibinfo{title}{Association of clinician diagnostic performance with machine learning--based decision support systems: a systematic review}}.
\newblock {\emph{\JournalTitle{JAMA network open}}} \textbf{\bibinfo{volume}{4}}, \bibinfo{pages}{e211276--e211276} (\bibinfo{year}{2021}).

\bibitem{lehman2015diagnostic}
\bibinfo{author}{Lehman, C.~D.} \emph{et~al.}
\newblock \bibinfo{journal}{\bibinfo{title}{Diagnostic accuracy of digital screening mammography with and without computer-aided detection}}.
\newblock {\emph{\JournalTitle{JAMA internal medicine}}} \textbf{\bibinfo{volume}{175}}, \bibinfo{pages}{1828--1837} (\bibinfo{year}{2015}).

\bibitem{gaube2021ai}
\bibinfo{author}{Gaube, S.} \emph{et~al.}
\newblock \bibinfo{journal}{\bibinfo{title}{Do as ai say: susceptibility in deployment of clinical decision-aids}}.
\newblock {\emph{\JournalTitle{NPJ digital medicine}}} \textbf{\bibinfo{volume}{4}}, \bibinfo{pages}{31} (\bibinfo{year}{2021}).

\bibitem{ruamviboonsuk2019deep}
\bibinfo{author}{Ruamviboonsuk, P.} \emph{et~al.}
\newblock \bibinfo{journal}{\bibinfo{title}{Deep learning versus human graders for classifying diabetic retinopathy severity in a nationwide screening program}}.
\newblock {\emph{\JournalTitle{NPJ digital medicine}}} \textbf{\bibinfo{volume}{2}}, \bibinfo{pages}{25} (\bibinfo{year}{2019}).

\bibitem{goodfellow2014explaining}
\bibinfo{author}{Goodfellow, I.~J.}
\newblock \bibinfo{journal}{\bibinfo{title}{Explaining and harnessing adversarial examples}}.
\newblock {\emph{\JournalTitle{arXiv preprint arXiv:1412.6572}}}  (\bibinfo{year}{2014}).

\bibitem{tinycancer}
\bibinfo{author}{Kleinman, Z.}
\newblock \bibinfo{title}{{NHS AI test spots tiny cancers missed by doctors}}.
\newblock \bibinfo{howpublished}{\url{https://www.bbc.co.uk/news/technology-68607059}} (\bibinfo{year}{2024}).
\newblock \bibinfo{note}{Accessed: 2024-09-17}.

\bibitem{dyer2021diagnosis}
\bibinfo{author}{Dyer, T.} \emph{et~al.}
\newblock \bibinfo{journal}{\bibinfo{title}{Diagnosis of normal chest radiographs using an autonomous deep-learning algorithm}}.
\newblock {\emph{\JournalTitle{Clinical radiology}}} \textbf{\bibinfo{volume}{76}}, \bibinfo{pages}{473--e9} (\bibinfo{year}{2021}).

\bibitem{leibig2022combining}
\bibinfo{author}{Leibig, C.} \emph{et~al.}
\newblock \bibinfo{journal}{\bibinfo{title}{Combining the strengths of radiologists and ai for breast cancer screening: a retrospective analysis}}.
\newblock {\emph{\JournalTitle{The Lancet Digital Health}}} \textbf{\bibinfo{volume}{4}}, \bibinfo{pages}{e507--e519} (\bibinfo{year}{2022}).

\bibitem{dvijotham2023enhancing}
\bibinfo{author}{Dvijotham, K.} \emph{et~al.}
\newblock \bibinfo{journal}{\bibinfo{title}{Enhancing the reliability and accuracy of ai-enabled diagnosis via complementarity-driven deferral to clinicians}}.
\newblock {\emph{\JournalTitle{Nature Medicine}}} \textbf{\bibinfo{volume}{29}}, \bibinfo{pages}{1814--1820} (\bibinfo{year}{2023}).

\bibitem{smith2023real}
\bibinfo{author}{Smith, J.}, \bibinfo{author}{Morgan, T.~N.}, \bibinfo{author}{Williams, P.}, \bibinfo{author}{Malik, Q.} \& \bibinfo{author}{Rasalingham, S.}
\newblock \bibinfo{journal}{\bibinfo{title}{Real-world performance of autonomously reporting normal chest radiographs in nhs trusts using a deep-learning algorithm on the gp pathway}}.
\newblock {\emph{\JournalTitle{arXiv preprint arXiv:2306.16115}}}  (\bibinfo{year}{2023}).

\bibitem{florkowski2008sensitivity}
\bibinfo{author}{Florkowski, C.~M.}
\newblock \bibinfo{journal}{\bibinfo{title}{Sensitivity, specificity, receiver-operating characteristic (roc) curves and likelihood ratios: communicating the performance of diagnostic tests}}.
\newblock {\emph{\JournalTitle{The Clinical Biochemist Reviews}}} \textbf{\bibinfo{volume}{29}}, \bibinfo{pages}{S83} (\bibinfo{year}{2008}).

\bibitem{nguyen2015deep}
\bibinfo{author}{Nguyen, A.}, \bibinfo{author}{Yosinski, J.} \& \bibinfo{author}{Clune, J.}
\newblock \bibinfo{title}{Deep neural networks are easily fooled: High confidence predictions for unrecognizable images}.
\newblock In \emph{\bibinfo{booktitle}{Proceedings of the IEEE conference on computer vision and pattern recognition}}, \bibinfo{pages}{427--436} (\bibinfo{year}{2015}).

\bibitem{lakshminarayanan2017simple}
\bibinfo{author}{Lakshminarayanan, B.}, \bibinfo{author}{Pritzel, A.} \& \bibinfo{author}{Blundell, C.}
\newblock \bibinfo{journal}{\bibinfo{title}{Simple and scalable predictive uncertainty estimation using deep ensembles}}.
\newblock {\emph{\JournalTitle{Advances in neural information processing systems}}} \textbf{\bibinfo{volume}{30}} (\bibinfo{year}{2017}).

\bibitem{blundell2015weight}
\bibinfo{author}{Blundell, C.}, \bibinfo{author}{Cornebise, J.}, \bibinfo{author}{Kavukcuoglu, K.} \& \bibinfo{author}{Wierstra, D.}
\newblock \bibinfo{title}{Weight uncertainty in neural network}.
\newblock In \emph{\bibinfo{booktitle}{International conference on machine learning}}, \bibinfo{pages}{1613--1622} (\bibinfo{organization}{PMLR}, \bibinfo{year}{2015}).

\bibitem{gal2016dropout}
\bibinfo{author}{Gal, Y.} \& \bibinfo{author}{Ghahramani, Z.}
\newblock \bibinfo{title}{Dropout as a bayesian approximation: Representing model uncertainty in deep learning}.
\newblock In \emph{\bibinfo{booktitle}{international conference on machine learning}}, \bibinfo{pages}{1050--1059} (\bibinfo{organization}{PMLR}, \bibinfo{year}{2016}).

\bibitem{kompa2021second}
\bibinfo{author}{Kompa, B.}, \bibinfo{author}{Snoek, J.} \& \bibinfo{author}{Beam, A.~L.}
\newblock \bibinfo{journal}{\bibinfo{title}{Second opinion needed: communicating uncertainty in medical machine learning}}.
\newblock {\emph{\JournalTitle{NPJ Digital Medicine}}} \textbf{\bibinfo{volume}{4}}, \bibinfo{pages}{4} (\bibinfo{year}{2021}).

\bibitem{gawlikowski2023survey}
\bibinfo{author}{Gawlikowski, J.} \emph{et~al.}
\newblock \bibinfo{journal}{\bibinfo{title}{A survey of uncertainty in deep neural networks}}.
\newblock {\emph{\JournalTitle{Artificial Intelligence Review}}} \textbf{\bibinfo{volume}{56}}, \bibinfo{pages}{1513--1589} (\bibinfo{year}{2023}).

\bibitem{hendrycks2016baseline}
\bibinfo{author}{Hendrycks, D.} \& \bibinfo{author}{Gimpel, K.}
\newblock \bibinfo{title}{A baseline for detecting misclassified and out-of-distribution examples in neural networks}.
\newblock In \emph{\bibinfo{booktitle}{In Proceedings of International Conference on Learning Representations (ICLR)}} (\bibinfo{year}{2017}).

\bibitem{da2021independent}
\bibinfo{author}{da~Silva, L.~M.} \emph{et~al.}
\newblock \bibinfo{journal}{\bibinfo{title}{Independent real-world application of a clinical-grade automated prostate cancer detection system}}.
\newblock {\emph{\JournalTitle{The Journal of pathology}}} \textbf{\bibinfo{volume}{254}}, \bibinfo{pages}{147--158} (\bibinfo{year}{2021}).

\bibitem{perincheri2021independent}
\bibinfo{author}{Perincheri, S.} \emph{et~al.}
\newblock \bibinfo{journal}{\bibinfo{title}{An independent assessment of an artificial intelligence system for prostate cancer detection shows strong diagnostic accuracy}}.
\newblock {\emph{\JournalTitle{Modern Pathology}}} \textbf{\bibinfo{volume}{34}}, \bibinfo{pages}{1588--1595} (\bibinfo{year}{2021}).

\bibitem{raciti2020novel}
\bibinfo{author}{Raciti, P.} \emph{et~al.}
\newblock \bibinfo{journal}{\bibinfo{title}{Novel artificial intelligence system increases the detection of prostate cancer in whole slide images of core needle biopsies}}.
\newblock {\emph{\JournalTitle{Modern Pathology}}} \textbf{\bibinfo{volume}{33}}, \bibinfo{pages}{2058--2066} (\bibinfo{year}{2020}).

\bibitem{evans2024understanding}
\bibinfo{author}{Evans, H.} \& \bibinfo{author}{Snead, D.}
\newblock \bibinfo{journal}{\bibinfo{title}{Understanding the errors made by artificial intelligence algorithms in histopathology in terms of patient impact}}.
\newblock {\emph{\JournalTitle{NPJ Digital Medicine}}} \textbf{\bibinfo{volume}{7}}, \bibinfo{pages}{89} (\bibinfo{year}{2024}).

\bibitem{azam2024digital}
\bibinfo{author}{Azam, A.~S.} \emph{et~al.}
\newblock \bibinfo{journal}{\bibinfo{title}{Digital pathology for reporting histopathology samples, including cancer screening samples--definitive evidence from a multisite study}}.
\newblock {\emph{\JournalTitle{Histopathology}}} \textbf{\bibinfo{volume}{84}}, \bibinfo{pages}{847--862} (\bibinfo{year}{2024}).

\bibitem{graham2023screening}
\bibinfo{author}{Graham, S.} \emph{et~al.}
\newblock \bibinfo{journal}{\bibinfo{title}{Screening of normal endoscopic large bowel biopsies with interpretable graph learning: a retrospective study}}.
\newblock {\emph{\JournalTitle{Gut}}} \textbf{\bibinfo{volume}{72}}, \bibinfo{pages}{1709--1721} (\bibinfo{year}{2023}).

\bibitem{park2023ai}
\bibinfo{author}{Park, C.~M.}
\newblock \bibinfo{title}{{AI: Workload Reduction by Autonomous Reporting of Normal Chest Radiographs}} (\bibinfo{year}{2023}).

\bibitem{lawton2024clinicians}
\bibinfo{author}{Lawton, T.} \emph{et~al.}
\newblock \bibinfo{journal}{\bibinfo{title}{Clinicians risk becoming ‘liability sinks’ for artificial intelligence}}.
\newblock {\emph{\JournalTitle{Future Healthcare Journal}}} \textbf{\bibinfo{volume}{11}} (\bibinfo{year}{2024}).

\bibitem{smith2020artificial}
\bibinfo{author}{Smith, H.} \& \bibinfo{author}{Fotheringham, K.}
\newblock \bibinfo{journal}{\bibinfo{title}{Artificial intelligence in clinical decision-making: rethinking liability}}.
\newblock {\emph{\JournalTitle{Medical Law International}}} \textbf{\bibinfo{volume}{20}}, \bibinfo{pages}{131--154} (\bibinfo{year}{2020}).

\bibitem{fdaapprove}
\bibinfo{author}{{U.S. Food \& Drug}}.
\newblock \bibinfo{title}{{Artificial Intelligence and Machine Learning (AI/ML)-Enabled Medical Devices}}.
\newblock \bibinfo{howpublished}{\url{https://www.fda.gov/medical-devices/software-medical-device-samd/artificial-intelligence-and-machine-learning-aiml-enabled-medical-devices}} (\bibinfo{year}{2024}).
\newblock \bibinfo{note}{Accessed: 2024-09-17}.

\bibitem{fdaguidance}
\bibinfo{author}{{U.S. Food \& Drug}}.
\newblock \bibinfo{title}{{Predetermined Change Control Plans for Medical Devices}}.
\newblock \bibinfo{howpublished}{\url{https://www.fda.gov/regulatory-information/search-fda-guidance-documents/predetermined-change-control-plans-medical-devicess}} (\bibinfo{year}{2024}).
\newblock \bibinfo{note}{Accessed: 2025-01-13}.

\end{thebibliography}

%To include, in this order: \textbf{Accession codes} (where applicable); \textbf{Competing interests} (mandatory statement). 

%The corresponding author is responsible for submitting a \href{http://www.nature.com/srep/policies/index.html#competing}{competing interests statement} on behalf of all authors of the paper. This statement must be included in the submitted article file.

\end{document}